\documentclass[journal]{IEEEtran}
\usepackage{amsmath}%
\usepackage{amsfonts}%
\usepackage{amssymb}%
\usepackage{multirow}
\usepackage{rotating}
\usepackage{graphicx}
\usepackage{subfigure}
\begin{document}
\title{Statistical Learning in Automated Troubleshooting: Application to LTE Interference Mitigation}
\author{\authorblockN{Moazzam Islam Tiwana, Berna Sayrac and Zwi Altman, \textit{IEEE Senior Member}}\\
\thanks{\authorrefmark{1}Copyright (c) 2010 IEEE. Personal use of this material is permitted. However, permission to use this material for any other purposes must be obtained from the IEEE by sending a request to pubs-permissions@ieee.org.}
\authorblockA{Orange Labs\\
 38-40 rue du G\'en\'eral Leclerc, 92130 Issy-les-Moulineaux, France \\
$\lbrace$moazzamislam.tiwana,berna.sayrac,zwi.altman\}@orange-ftgroup.com
}
}
\maketitle
\begin{abstract}
This paper presents a method for automated healing as part of off-line automated troubleshooting.
The method combines statistical learning with constraint optimization.
The automated healing aims at locally optimizing radio resource management (RRM) or system parameters of cells with poor performance in an iterative manner.
The statistical learning processes the data using Logistic Regression (LR) to extract closed form (functional) relations between Key Performance Indicators (KPIs) and Radio Resource Management (RRM) parameters.
These functional relations are then processed by an optimization engine which proposes new parameter values.
The advantage of the proposed formulation is the small number of iterations required by the automated healing method to converge, making it suitable for off-line implementation.
The proposed method is applied to heal an Inter-Cell Interference Coordination (ICIC) process in a 3G Long Term Evolution (LTE) network which is based on soft-frequency reuse scheme.
Numerical simulations illustrate the benefits of the proposed approach.
\end{abstract}
\section{Introduction}
Efficient management of future Beyond 3G and 4G networks is a major challenge for network operators \cite{ref_1}. The wireless ecosystem is becoming more and more heterogeneous with co-existing/co-operating technologies and deployment scenarios (i.e. macro, micro, pico and femto cell structures). Fault management or troubleshooting is an important building block of network operation.
Troubleshooting comprises three functionalities: fault detection
(i.e. detecting failures or poor performance as soon as they occur); fault diagnosis
(i.e. determining the cause of failure or of poor performance), and fault recovery or
healing (i.e. repairing the problem) \cite{ref_2}.
\\
The importance of efficient fault management has motivated the development of automated methods and tools for diagnosis and healing.
In this work, the main focus is given to automated healing.
It is supposed that a given cell with poor performance has been identified (fault detection) and the cause of the degraded performance has been diagnosed as a bad setting of a specific Radio Resource Management (RRM) or a system parameter (fault diagnosis)~\cite{ref_2}\cite{ref_3}.
The automated healing process aims at locally optimizing the value of this parameter, taking into account the Key Performance Indicators (KPIs) of the faulty cell and those of its neighbours.
In other words, the RRM parameter that is found as the fault cause by the diagnosis process is optimized by the automated healing module.
\\
Local type of optimization or "steered optimization" has been studied in the literature, based on combinatorial optimization in conjunction with the interference matrix to tackle local problems detected in the network \cite{ref_4}. 
This approach uses a network simulator and can be implemented as an advanced functionality of a cell planning tool.
The focus of this paper is to develop an automated healing method based on measurements.
More precisely, our aim is to conceive an automated healing method that uses statistical learning of measured data and constraint optimization.
The method is denoted as Statistical Learning Automated Healing (SLAH).
The SLAH module can be located at the management plane, e.g. in the Operation and Maintenance Centre (OMC) where abundant data is available.
The method is iterative with a time resolution of a day, and should therefore converge in a few iterations.
To achieve this requirement, a statistical learning approach using Logistic Regression (LR) is proposed that extracts the functional relations between KPIs and RRM parameters and comprises the \emph{statistical model}.
It is noted that the data is noisy due to the random character of the traffic and of the radio channel, but also due to imprecisions of the measurements.
After each iteration, the statistical model is updated using the additional data and its precision is improved.
The model is then introduced into the optimization engine and is processed directly to derive the next RRM parameter.
This approach has the merit of converging rapidly.
The performance of the SLAH is evaulated on an interference mitigation use case, namely an Inter-Cell Interference 
Coordination (ICIC) problem in a LTE network.The choice of this use case is motivated by the importance of interference 
mitigation in OFDMA (LTE/WiMax) networks, since it allows to improve the system performance, and particularly, to reach 
the strict requirements for cell edge bit rates (defined in B3G and 4G network standards, such as \cite{requirements}). 
In this context, ICIC is one of the efficient approaches to mitigate interference.
Interference mitigation techniques such as ICIC can considerably improve Signal to
Interference plus Noise (SINR) and hence bit rates, particularly at cell edge. As a result,
better network performance and user Quality of Service (QoS) are achieved,
including reduced File Transfer Time (FTT), Block Call Rate (BCR) and Drop Call Rate (DCR).
Different interference mitigation methods have been proposed for OFDMA systems,
such as fractional reuse and soft reuse schemes \cite{ref_10}-\cite{R_Y_Chang}.
When different power allocation for the mobile users is associated with different portions of
the frequency bandwidth, the frequency reuse is called a \emph{soft reuse} scheme, and will be
considered here in the context of automated healing.
\\
The paper is organized as follows: Section II introduces the concept and the system model for the SLAH and explains its different building blocks. Section III describes the LTE ICIC model that is used in the SLAH case study.
The adaptation of the SLAH to heal the ICIC process is developed in Section IV.
Numerical results are presented in Section V followed by concluding remarks in Section VI.
\section{System Model for Automated Healing}
\begin{figure}[!hbtp]
\centering
\includegraphics[width=90mm,height=50mm]{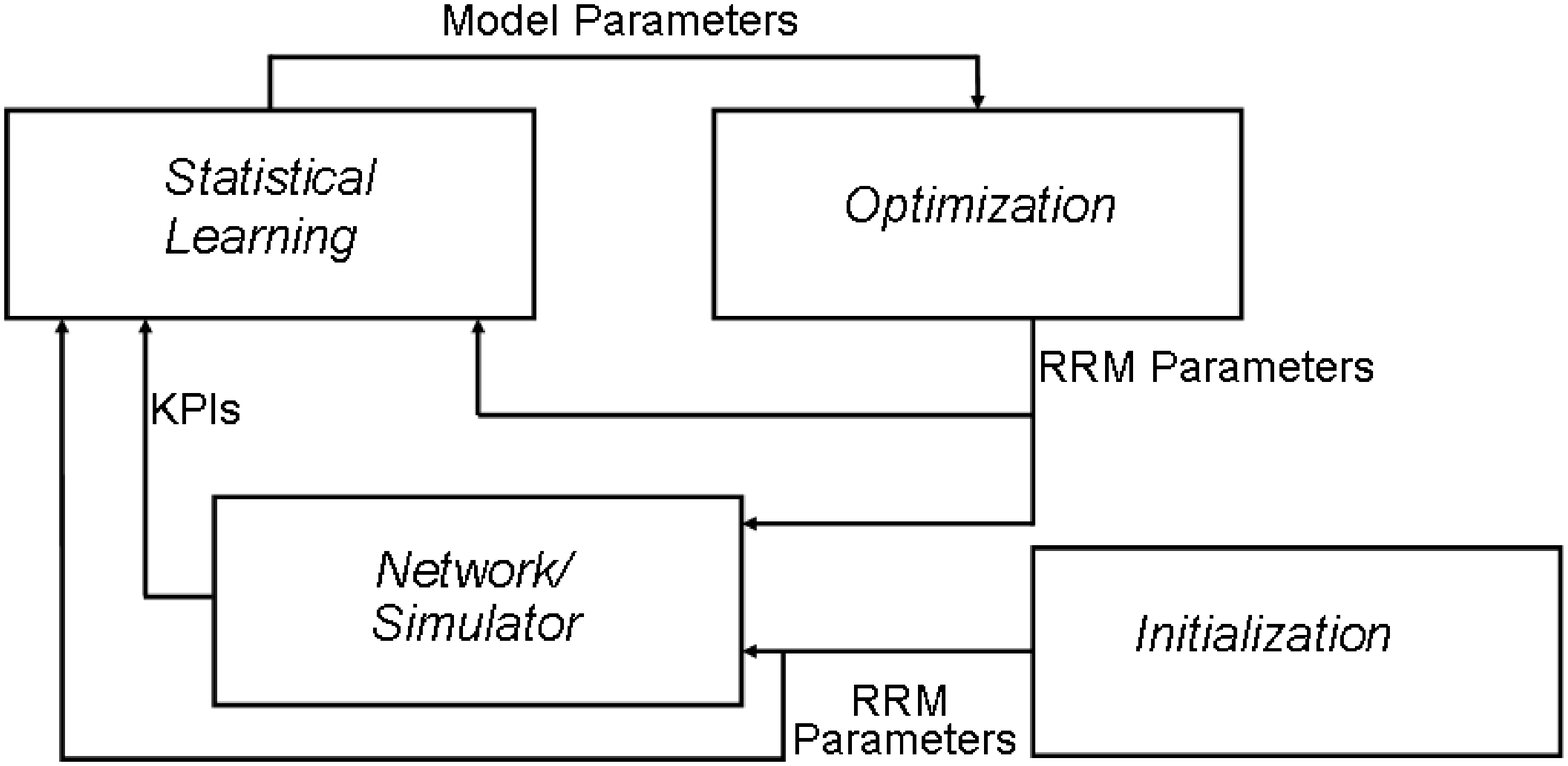}
\caption{SLAH block diagram.}
\label{fig:figure1}
\end{figure}

It is assumed that the fault cause has been diagnosed as a specific RRM parameter (such as handover/mobility, admission and congestion control thresholds) whose value has degraded the performance of the eNodeB (eNB).
An example of such a case is presented in \cite{ref_3} where the bad setting of the add/drop window of a NodeB in a UMTS system is diagnosed.
The purpose of the SLAH is to iteratively optimize the value of this RRM parameter using local information from the eNB and its neighbors.
Hence, the automated healing is a \emph{local optimization} process.
The SLAH block diagram is presented in Figure \ref{fig:figure1}.
The system model comprises four blocks:
\\\emph{Initialization block:}
The Initialization block provides the initial RRM parameter to the faulty eNB in the Network/Simulator block and to the Statistical Learning block.
\\\emph{Network/Simulator block:}
The Network/Simulator block represents the real network or the network simulator.
It measures (case of real network) or calculates (case of network simulator) a set of KPIs of an eNB and of its neighbors for each new RRM parameter introduced by the Initialization or the Optimization block.
\\\emph{Statistical Learning block:}
The Statistical Learning block extracts the functional relations, known as the \emph{statistical model}, between the KPIs and the RRM parameter through Logistic Regression (LR)~\cite{reg_book2}.
LR fits the data into the functional form denoted as $logistic \mbox{ } function$: $f_{log}(z)=\frac{1}{1+exp^{-z}}$.
The $f_{log}(z)$ can describe saturation effects at its extremities as often encountered in KPIs in communication networks.
\\
Let $y_{m,i}$ denote the $i^{th}$ sample value of the $m^{th}$ dependent variable $y_m$ (i.e. the $m^{th}$ KPI) corresponding to the $i^{th}$ sample value $x_i$ of the explanatory variable $x$ (i.e. the RRM parameter).
LR models $y_{m,i}$ as follows:
%
\begin{equation}
\label{eq:gen_mod_eq}
y_{m,i}=f_{log}(\eta_{m,i})+\epsilon_i
\end{equation}
where $\eta_{m,i}=\beta_{m,0}+\beta_{m,1}x_i$ is the linear predictor representing the contribution of the explanatory variable sample $x_i$, and $\epsilon_i$ is the the residual error.
The $\beta$s are the regression coefficients whose values have to be estimated using maximum likelihood estimation~\cite{gen_lin}.
Hence from~\eqref{eq:gen_mod_eq}, the functional relation between $\hat{y}$, i.e., $y$ estimated by LR, and $x$ can be written as:
\begin{equation}
\hat{y}_m(x)=f_{log}(\beta_{m,0}+\beta_{m,1}x) 
\label{eq:y_x_funrel}
\end{equation}
%
\\\emph{Optimization block:}
The Optimization block calculates the optimal RRM value using the current statistical model.
It determines $\hat{x}$, i.e., the value of the RRM parameter $x$ that minimizes a cost function of a set of KPIs denoted as the \emph{optimization set} $A_o$, subject to constraints on a second set of KPIs denoted as the \emph{constraint set} $A_c$.
Considering that $\hat{y}_m(x)$ has the functional relation form as in~\eqref{eq:y_x_funrel}, the optimization problem can be formulated as:
\begin{equation}
\hat{x}=argmin_{x} C(x)
\label{eq:opt_prob}
\end{equation}
where $C=\sum_{m \in A_o} w_m \hat{y}_m(x)$ is the cost function and $w_m$ is the weight given to $\hat{y}_m(x)$.
\\
\\
The automated healing process is iterative.
At each iteration, a new RRM parameter value is proposed (by the Initialization block during the initialization iterations and by the Optimization block during the optimization iterations) to update the RRM setting of the faulty eNB in the Network/Simulator block.
The performance of the faulty eNB and of its neighbors with this new RRM value is assessed by the Network/Simulator block through a set of KPI values obtained at the end of the measurement period (typically one day).
Thus, a data point comprising a RRM parameter value and the corresponding KPIs is obtained.
This data point together with the previously obtained data points are used by the Statistical Learning block to refine the statistical model which is then used by the Optimization block to generate the RRM parameter value of the next iteration.
Thus, as the iterations progress, on the average, the model precision improves and is used by the Optimization block to find a better value for the RRM parameter.
\section{System Model for Interference Mitigation}
\begin{figure}[!hbtp]
\centering
\includegraphics[height=90mm,width=80mm]{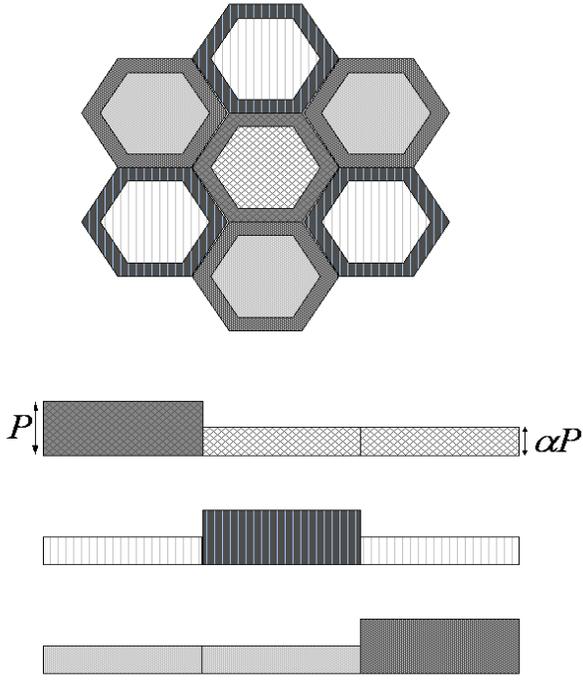}
\caption{System Model.}
\label{Figure:figure2}
\end{figure}

%
The performance of the proposed automated healing method is evaluated on an ICIC scheme which uses soft-frequency reuse.
%
Consider a downlink ICIC scheme that combines two resource allocation mechanisms: Physical Resource Block (PRB) allocation to frequency subbands and coordinated power allocation.
In the soft-reuse one scheme, the total available bandwidth is reused in all the cells while the transmitted power for a portion of the bandwidth of a cell can be adapted to solve interference related QoS problems.
Figure \ref{Figure:figure2} 
presents the power-frequency allocation model in a seven adjacent cell layout.
The frequency band is divided into three disjoint subbands.
One subband is allocated to mobiles with the worst signal quality and is denoted interchangeably as a \emph{protected band} or as an \emph{edge band} with transmit power $P$.
A user with poor radio conditions is often situated at the cell edge, but could also be closer to the base station and experience deep shadow fading.
The remaining two frequency subbands are denoted as \emph{centre bands} with transmit power reduced by a factor $\alpha$, namely $\alpha P$.
The interference produced by an eNB to its neighbours can be controlled by the parameter $\alpha$ of this eNB.
The main downlink interference in the system originates from eNB transmissions on the centre band (to centre cell users) which interfere with neighboring cell edge users utilizing their edge (protected) band.
When an eNB strongly interferes with the users of its neighbours, the ICIC mechanism allows to reduce the transmission power for the centre band.
\\
Resource block allocation is performed based on a priority scheme for accessing the protected subbands.
A quality metric $q_u$ is calculated using pilot channel signal strengths as
 $q_u=  \frac{Pr_{su}}{\mathop{\sum} \limits_{j \neq s}{ Pr_{ju} }+ \sigma_z^2}$.
Here $s$ stands for the serving eNB of user $u$, $Pr_{ju}$ denotes the mean pilot power received by the user $u$ of a signal transmitted by the eNB $j$, and $\sigma_z^2$ is the noise power spectral density.
$q_u$ is similar to the SINR with the difference that in the present ICIC scheme, the data channels used to calculate the SINR are subject to power control.
The $q_u$ metric is calculated for all users which are then sorted according to this metric.
Users with the worst $q_u$ are allocated resources from the protected band and benefit from maximal transmission power of the eNB.
When the protected subband is full, the resource block allocation continues from the centre band.
\\
Note that the soft-reuse ICIC scheme is characterized by two other parameters in addition to
$\alpha$: 1) the number of PRBs assigned to the center and edge bands; 2) the threshold that determines the boundary between center and cell edge users.
In this work, for reasons of simplicity, we deal only with one parameter.
The proposed algorithm can easily be generalized to multiple RRM case, however with an increased complexity.
The choice of the $\alpha$ parameter is motivated by the simplicity in its implementation, which is carried out by a simple power control on a pre-defined set of subcarriers while the other two parameters require modifications of the PRB scheduling strategy.
Nevertheless, the proposed algorithm is equally applicable to the other two parameters without any major alterations.
\section{SLAH for Interference Mitigation}
This section describes the adaptation of the SLAH to interference mitigation in a LTE network by locally optimizing the parameter $\alpha$ of the interfering eNBs.
Denote by $eNB_c$ ($c$ standing for $\it{central}$) an eNB with degraded performance.
It is assumed that the cause of the degraded performance has been diagnosed and is related to excessive inter-cell interference which can be effectively mitigated by a soft-reuse ICIC scheme.
The first tier neighbours of $eNB_c$ are denoted by $eNB_j$, $j\in NS1$ where $NS1$ is the index set of the first-tier neighbours of $eNB_c$.
The specificity of the interference mitigation use case is the following: to heal $eNB_c$, the parameters $\alpha_j$ of $eNB_j$, $j\in NS1$ , are optimized, while  $\alpha_c$ of $eNB_c$ remains unchanged.
\\
We use the notion of \emph{coupling} between $eNB_j$ and $eNB_c$ which is expressed in terms of the interference that $eNB_j$ produces on the users connected to $eNB_c$ and can be written in terms of the downlink interference matrix element $I_{cj}$~\cite{ref_4}.
Hence the bigger the $I_{cj}$, the stronger the coupling between the two eNBs.
Note that the matrix element $I_{cj}$ is equal to the time average of the sum of interferences perceived by the mobiles attached to $eNB_c$ and generated by downlink transmissions to the mobiles of $eNB_j$.
Denote by $s$, $s\in NS1$, the index of the eNB which is the most coupled with $eNB_c$, namely $s=argmax_{j}(I_{cj})$, $j\in NS1$.
To reduce the complexity of the SLAH process, we propose to adjust the $\alpha_j$ parameter according to the degree of coupling between $eNB_j$ and $eNB_c$.
Hence, we define a functional relation between $\alpha_s$ and $\alpha_j$ that accounts for the coupling mentioned above:
\begin{equation}
\label{eq:alphaformulasimple}
\alpha_j=\alpha_s+(1-\alpha_s)(1-\frac{I_{cj}}{I_{ci}}).
\end{equation}
Note that the smaller the coupling between $eNB_j$ and $eNB_c$, the lesser the power reduction applied to $eNB_j$.
Thus, by using~\eqref{eq:alphaformulasimple}, only $\alpha_s$ needs to be optimized instead of all the first tier $\alpha_j$s.
The self-healing process can be performed simultaneously on any number of eNBs provided they are not direct neighbours.
\\
The SLAH aims at minimizing the FTT for $eNB_c$ and of its first-tier neighbours while verifying constraints on their BCRs ($BCR_j$, $j\in c\cup NS1$).
We define the cost function for the optimization as follows:
\begin{equation}
\label{eq:gen_opt}
C= FTT_c+\sum_{j\in NS1} \omega_j FTT_j.
\end{equation}
It is noted that $FTT_j$ is a function of $\alpha_j$ and hence, via equation~\eqref{eq:alphaformulasimple}, of $\alpha_s$.
The weighting coefficients $\omega_j$ depend on the relative contribution of $I_{cj}$ with respect to the sum on all eNBs in $NS1$ and are given by:
\begin{equation}
\label{eq:omega_coeff}
\omega_j=\frac{I_{cj}}{\sum_{l \in NS1}I_{cl}}
\end{equation}
satisfying the condition $\sum_{j\in NS1}\omega_j=1$.
The optimization problem can now be formulated as follows:
\begin{equation}
\label{eq:gen_opt_simp}
\alpha_s=argmin_{\alpha_s'}C(\alpha_s')
\end{equation}
subject to $BCR_{j}<BCR_{th} \mbox{ };\mbox{ } j\in c \cup NS1$.
$BCR_{th}$ is the threshold for $BCR_j$ above which communication quality is unacceptably poor.
The $FTT$ and $BCR$ indicators in equations~\eqref{eq:gen_opt} and~\eqref{eq:gen_opt_simp}, are given in the form of LR function~\eqref{eq:y_x_funrel} obtained using the LR.
\\
The SLAH can be further improved by introducing a generalized interference matrix element $I_{cj}'$ in equation~\eqref{eq:alphaformulasimple} by introducing an additional KPI, namely the BCR:
\begin{equation}
\label{eq:newint}
I_{cj}'=I_{cj}e^{-\gamma B_j} \mbox{ };\mbox{ } B_j=BCR_j/max_{l\in NS1}(BCR_l).
\end{equation}
One can see that the higher the $B_j$ (i.e. the normalized $BCR_j$), the smaller the $I_{cj}'$ and consequently, the smaller the modification of $\alpha_j$.
The significance of equation~(\ref{eq:newint}) is that, in order to improve the performance of $eNB_c$, the decrease made in $\alpha_j$ due to $I_{cj}$ is limited by the degradation in $BCR_j$.
Note that the constant $\gamma$ allows tuning the effect of $B_j$.
\\
Denote a $\it{data\mbox{ }point}$ $p_k^j$ as the vector
$p_k^j=(\alpha_j,FTT_j,BCR_j)_k$, where $j \in c\cup NS1$ and $k$ denotes the iteration index.
Since the SLAH starts with an initial data point and generates a new data point at each iteration, 
the iteration index equals the total number of generated data points.
The set of $k$ data points for $eNB_j$, $j\in c\cup NS1$, is denoted by $P_k^j$. 
The SLAH algorithm is given in Table~\ref{tab:SLAH}.
\begin{table}[]
\centering
\begin{tabular}{|l|}\hline
\it{Initialization:}\\
1. Identify the most coupled eNB $eNB_s$ with $eNB_c$ among the\\ neighbours in $NS1$\\
2. Generate the initial set of $k$ data points $P_k^j$, $j\in c\cup NS1$, by applying\\ $k$ different $\alpha_s$ values
(together with the associated $\alpha_j$ values) to the\\ network/simulator one by one and obtaining the corresponding KPIs.\\
\it{Repeat until convergence:} \\
3. For each $eNB_j$, compute the statistical model using LR for $FTT$ and\\ $BCR$ using the corresponding data points in $P_k^j$\\
4. Compute a new $\alpha$ vector containing the new values of $\alpha_j$, $j\in NS1$\\ (using equations~\eqref{eq:alphaformulasimple} and~\eqref{eq:gen_opt_simp})\\
5. Apply the new $\alpha_j$ values to the network/simulator and observe ($FTT_j$)\\ and ($BCR_j$), $j\in c\cup NS1$.
Compute the new data point $p_{k+1}^j$\\ 
6. Update $P_{k+1}^j$: $P_{k+1}^j=P_k^j\cup p_{k+1}^j$\\
7. k=k+1\\
\it{End Repeat}\\\hline
\end{tabular}
\caption{The complete SLAH Algorithm}
\label{tab:SLAH}
\end{table}
\section{Case Study}
\subsection{Simulation Scenario}
A LTE network comprising 45 eNBs in a dense urban environment and having bandwidth of 5MHz is simulated using the MATLAB simulator described in \cite{ridha}.
The simulator performs correlated Monte Carlo snapshots with a time resolution of 1 second to account for the time evolution of the network.
FTP traffic with a file size of 6300 Kbits is considered.
Call arrivals are generated using the Poisson process and the communication duration of each user depends on its bit rate.
The maximum number of PRBs in an eNB, i.e. the capacity, is fixed to 24 PRBs with 8 PRBs in each sub-band.
The number of PRBs that can be allocated to a user can vary from 1 to 4, allocated on the first-come first-served basis.
No mobility is taken into account.

For each new value of $\alpha$, the simulator runs for 2500 time steps (seconds) to allow the convergence of the processed KPIs.
The BCR and FTT KPIs used by the SLAH algorithm are averaged on an interval varying from 500 to 2500 seconds while discarding the samples of first 500 seconds during which the network reaches a steady state.
It is noted that for a given traffic demand, the BCR provides a capacity indicator while the FTT is more related to the user 
perceived QoS.
The simulated LTE system includes a simple admission control process based on signal strength: A simple admission control has been implemented based on signal strength. A mobile selects the eNB with the highest Reference Signal Received Power (RSRP) and is admitted if it is above -104 dBm and if at least one PRB is available. The mobile throughput is calculated from SINR using quality tables obtained from link level simulations.
The SINR and consequently the bit rate of a mobile are updated after each simulation time step. 
The interference matrix elements used in equations~\eqref{eq:alphaformulasimple},~\eqref{eq:omega_coeff} and~\eqref{eq:newint} are calculated only once for the reference solution (see paragraph below) during a longer time interval varying from 500 to 7000 seconds to achieve accurate average results.
\subsubsection*{Reference Solution}
An optimal default value for $\alpha$, known as \textit{reference solution}, is calculated as 0.5 for all eNBs in the network. 
The default $\alpha$ value is determined by varying it simultaneously for all eNBs from 0.0125 to 1 in steps of 0.0125.
For each $\alpha$, the network performance is assessed in terms of the mean BCR and mean FTT.
The minimum values for both BCR and FTT are obtained in the $\alpha$ interval $[0.5,0.7]$.
The value of $\alpha=0.5$ is selected as the default value due to the smaller inter-cellular interference and the minimum energy consumption in the network.
\subsection{Automated Healing Scenario}
A problematic eNB with the worst performance in the simulated network (in terms of BCR and FTT), namely $eNB_{c=13}$, is selected for automated healing using the SLAH algorithm.
The $eNB_j$, where $j\in NS1=\left\{14,15,22,23,43,45\right\}$, is one of the six first tier neighbours of $eNB_{c=13}$.
$\alpha_c$ is fixed to the reference default value of 0.5.
The index set $NS2$ of the second tier neighbours of the problematic eNB consists of $NS2=\left\{1,10,11,16,18,24,37,44\right\}$.
Denote by \textit{optimization zone} the subnetwork comprising $eNB_{c=13}$ and its first tier neighbours $NS1$, and by \textit{evaluation zone} the subnetwork comprising $eNB_{c=13}$ and its first two tier neighbours $NS1$ and $NS2$.
The $eNB_{s=45}$ is the eNB most coupled with $eNB_{c=13}$.
\subsection{Results}
The SLAH algorithm is applied using the generalized interference matrix~\eqref{eq:newint} in~\eqref{eq:alphaformulasimple},
with $\gamma$=-0.3.
The first five values of $\alpha_{s=45}$ in Table \ref{tab:alpha} are chosen for the initialization phase (Phase-I in the Table) of the SLAH.
The next seven values are calculated iteratively by the SLAH algorithm during the optimization phase (Phase-II in the Table).
The values of $\alpha_{j=14}$, $\alpha_{j=15}$, $\alpha_{j=22}$, $\alpha_{j=23}$ and $\alpha_{j=43}$ are calculated using equation~\eqref{eq:alphaformulasimple}.
In spite of the inherent noise present in the generated data, one can see from the values depicted in Phase-II that $\alpha_{s=45}$ converges in a few iterations.
$\alpha_{s=45}=0.46$ is chosen as the optimized solution.

\begin{figure}[htp]
\centering
\subfigure[]{\label{fig:figure3a}\includegraphics[width=84mm,height=61mm]{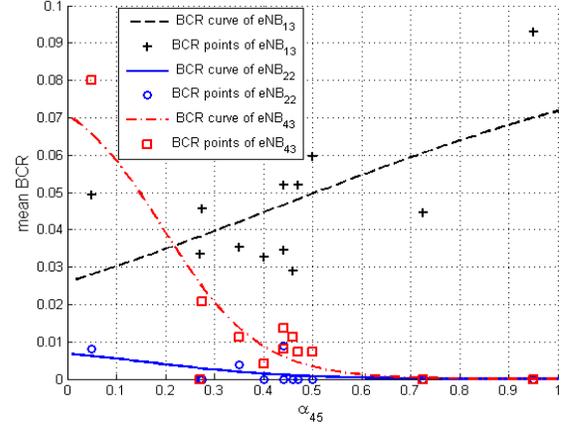}}
\subfigure[]{\label{fig:figure3b}\includegraphics[width=87mm,height=61mm]{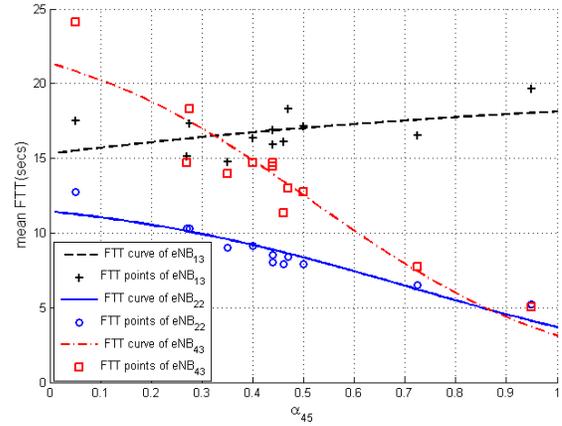}}
\caption{Mean KPI values and LR regression curves as a function of $\alpha_{s=45}$ for $eNB_{j=13}$, $eNB_{j=22}$ and $eNB_{j=43}$; mean BCR (a)  mean FTT (b).}
\end{figure}

\begin{figure}[htp]
\centering
\subfigure[]{\label{fig:figure4a}\includegraphics[width=84mm,height=55mm]{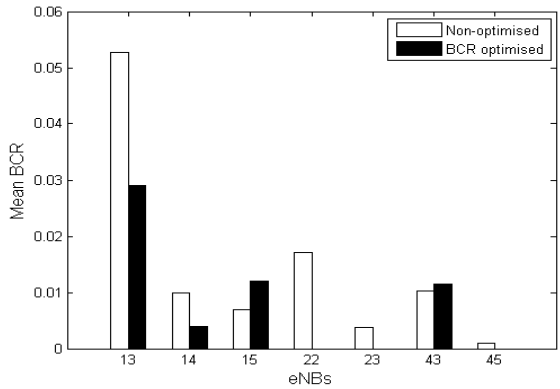}}
\subfigure[]{\label{fig:figure4b}\includegraphics[width=84mm,height=55mm]{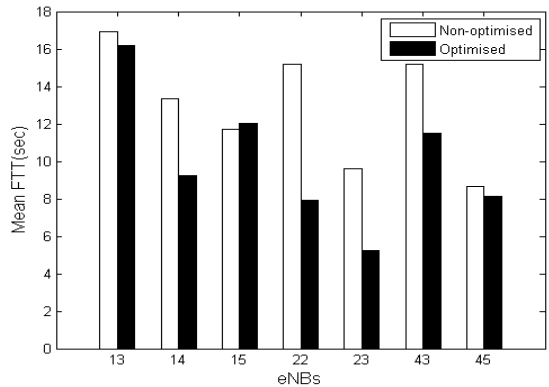}}
\caption{KPI of the eNBs in the optimization zone for the reference solution (white) and optimized (black) network conditions; mean BCR (a) and mean FTT (b).}
\end{figure}

\begin{figure}[htp]
\centering
\subfigure[]{\label{fig:figure5a}\includegraphics[width=90mm,height=52mm]{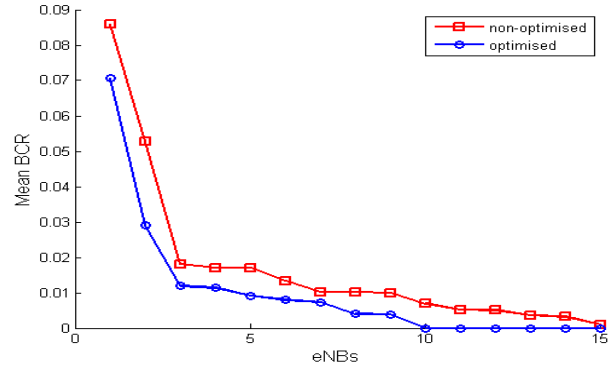}}
\subfigure[]{\label{fig:figure5b}\includegraphics[width=84mm,height=52mm]{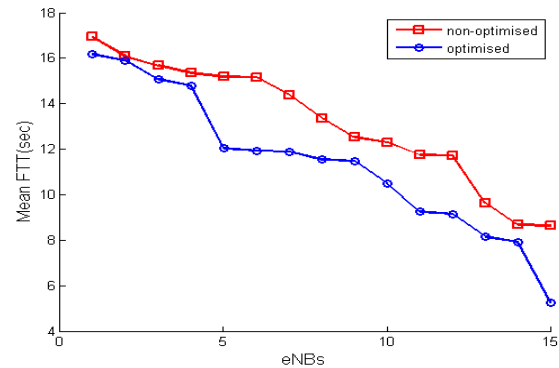}}
\caption{KPIs in descending order for the eNBs in the evaluation zone; mean BCR (a) and mean FTT (b).}
\end{figure}

\begin{table}{}
\centering
\begin{tabular}{|c|c|c|c|c|c|c|c|}
\hline
 & $\alpha_{c=13}$ &	$\alpha_{j=14}$	& $\alpha_{j=15}$	& $\alpha_{j=22}$ &	$\alpha_{j=23}$ & $\alpha_{j=43}$ &	$\alpha_{s=45}$\\
\cline{2-8}
\multirow{5}*{\rotatebox{90}{Phase I}} & 0.50 &	0.97	& 0.95	& 0.98 &	0.99 & 0.96 &	 0.95\\\cline{2-8}
 & 0.50	& 0.85 &	0.74	& 0.87 & 0.97	& 0.79 &	0.73\\\cline{2-8}
 & 0.50	& 0.74	& 0.53	& 0.77	& 0.94	& 0.61	& 0.50\\\cline{2-8}
 & 0.50	& 0.62	& 0.31	& 0.67	& 0.92	& 0.43	& 0.28\\\cline{2-8}
 & 0.50	& 0.50	& 0.10	& 0.57	& 0.89	& 0.26	& 0.05\\\hline
\multirow{7}*{\rotatebox{90}{Phase II}} & 0.50	& 0.61	& 0.31	& 0.67	& 0.92	& 0.43	& 0.27\\\cline{2-8}
 & 0.50	& 0.66	& 0.38	& 0.70	& 0.93	& 0.49	& 0.35\\\cline{2-8}
 & 0.50	& 0.68	& 0.43	& 0.73	& 0.93	& 0.53	& 0.40\\\cline{2-8}
 & 0.50	& 0.70	& 0.47	& 0.74	& 0.94	& 0.53	& 0.44\\\cline{2-8}
 & 0.50	& 0.70	& 0.47	& 0.74	& 0.94	& 0.56	& 0.44\\\cline{2-8}
 & 0.50	& 0.72	& 0.50	& 0.76	& 0.94	& 0.59	& 0.47\\\cline{2-8}
 & 0.50	& 0.72	& 0.49	& 0.75	& 0.94	& 0.58	& 0.46\\\cline{2-8}
\hline
\end{tabular}
 \caption{Phase-I shows the initially chosen $\alpha$ values. Phase-II shows the $\alpha$ values calculated during optimization.}
   \label{tab:alpha}
\end{table}
Figures \ref{fig:figure3a} and \ref{fig:figure3b} 
show the mean BCR and FTT data points respectively as a function of $\alpha_{s=45}$ together with the LR curves for $eNB_{c=13}$, $eNB_{j=22}$ and $eNB_{j=43}$ after convergence (at the end of the $7^{th}$ optimization iteration).
The KPI curves for $eNB_{j=14}$, $eNB_{j=15}$, $eNB_{j=23}$ and $eNB_{s=45}$ are not shown as they have a similar trend.
The concentration of KPI data points around $\alpha_{s=45}=0.45$ indicates the convergence of the SLAH algorithm.
\\
Figure \ref{fig:figure4a} 
shows the gain brought about by the SLAH algorithm for the optimization zone (set $NS1$ of eNBs).
The mean BCR of the problematic $eNB_{c=13}$ is reduced by 45\% with respect to the reference solution, from 5.28\% to 2.9\%.
The average improvement of the mean BCR in the first tier ($NS1$) is 44\% with respect to the reference solution.
%
\\
In the case of mean FTT, the improvement brought about by the SLAH algorithm in the optimized zone with respect to the reference solution is shown in Figure~\ref{fig:figure4b}.
The mean FTT of $eNB_{c=13}$ is reduced by 6.31\% and the average improvement of the mean FTT in the first tier is 26.6\%.
This improvement is related to the optimized interference management in the first tier of the problematic eNB.
The decrease in interferences improves the SINR values and consequently the bit rates and the FTT values.
Furthermore, the improvement in power resource allocation decreases the sojourn time of users that monopolize scarce radio resources and results in the improvement in~BCR.
%
\\
Figures \ref{fig:figure5a} and \ref{fig:figure5b}
show, in descending order, the mean BCR and the mean FTT respectively for the reference (square) and the optimized (circle) eNBs in the evaluation zone ($eNB_{c=13} \cup NS1 \cup NS2$).
It is noted that the order of the stations in the two curves of each Figure may not be preserved.
One can see that on the average, the mean BCR and mean FTT in the evaluation zone are improved.
The average improvement of FTT in the evaluation zone is of 13\%.

%
\section{Conclusion}
This paper has presented a new approach, the SLAH, for automated healing of cells with poor performance.
The SLAH is an iterative optimization algorithm that uses statistical learning in conjunction with a simple optimization module.
During each iteration, the RRM solution computed by optimization block is improved jointly with the improvement in the statistical model.
The SLAH can be implemented in the management plane, e.g. in the OMC in an off-line mode.
It has been successfully applied to heal a downlink ICIC parameter of an eNB with degraded performance due to excess downlink inter-cell interference in a LTE network.
The proposed approach has several attractive features: it is generic and can be easily adapted to deal with different types of faulty parameters; it performs well in the presence of noisy data; and it converges in a very small number of iterations.
The SLAH method provides the basis for designing self-healing algorithms.

%
%
%
%
%
%
%
%
%
%
\end{document}